%% file: cvpr.tex
\documentclass[10pt,twocolumn,letterpaper]{article}
\usepackage{cvpr}             
\usepackage{graphicx}
\usepackage{amsmath}
\usepackage{amssymb}
\usepackage{booktabs}
\usepackage{soul}

\usepackage[pagebackref,breaklinks,colorlinks]{hyperref}
\makeatletter
\@namedef{ver@everyshi.sty}{}
\makeatother
\usepackage{tikz}
\usepackage{fancyhdr}

\usepackage{tikz}
\usepackage{pgfplots}
\usetikzlibrary{patterns}
\pgfplotsset{compat=1.16}
\usepackage{wrapfig}
\usepackage{subcaption}
\usepackage{tikz}
\usepackage{pgfplots}
\usetikzlibrary{patterns}
\pgfplotsset{compat=1.16}
\usepackage{wrapfig}
\usepackage{subcaption}
\usepackage{smartdiagram}  
\usepackage[edges]{forest}
\usetikzlibrary{arrows.meta}
\usepackage{rotating}
\usetikzlibrary{shapes,arrows,positioning}

\usepackage{csquotes}
\usepackage{graphics}
\usepackage{color}
\usepackage{multicol}
\usepackage{multirow}
\usepackage{adjustbox}

\def\ourmodel{\textit{PASNet}}

\usepackage{hyperref}

\usepackage[capitalize]{cleveref}
\crefname{section}{Sec.}{Secs.}
\Crefname{section}{Section}{Sections}
\Crefname{table}{Table}{Tables}
\crefname{table}{Tab.}{Tabs.}

\def\cvprPaperID{14} 
\def\confName{CVPR}
\def\confYear{2022}

\begin{document}

\title{Pyramidal Attention for Saliency Detection}

\author{Tanveer Hussain$^{1}$,  Abbas Anwar$^{2}$, Saeed Anwar$^{3,4,5,6}$, Lars Petersson$^4$, Sung Wook Baik$^{1}$\thanks{Corresponding author: Sung Wook Baik (\textit{sbaik@sejong.ac.kr}), for quick correspondence: \textit{tanveer445@ieee.org}}\\
Sejong University$^1$, Abdul Wali Khan University$^2$,  Australian National University$^3$, \\Data61-CSIRO$^4$, University of Technology Sydney$^5$, University of Canberra$^6$
}
\maketitle

\input{sections/abs}

\input{sections/intro}
\input{sections/related}

\input{sections/methodology}
\input{sections/experiments}

\vspace{-3mm}
\input{sections/conclusion}

{\small
\bibliographystyle{cvpr}
\bibliography{main}
}

\end{document}

%% file: sections/abs.tex
\begin{abstract}
Salient object detection (SOD) extracts meaningful contents from an input image. RGB-based SOD methods lack the complementary depth clues; hence,  providing limited performance for complex scenarios. Similarly, RGB-D models process RGB and depth inputs, but the depth data availability during testing may hinder the model's practical applicability. This paper exploits only RGB images, estimates depth from RGB, and leverages the intermediate depth features. We employ a pyramidal attention structure to extract multi-level convolutional-transformer features to process initial stage representations and further enhance the subsequent ones. At each stage, the backbone transformer model produces global receptive fields and computing in parallel to attain fine-grained global predictions refined by our residual convolutional attention decoder for optimal saliency prediction. We report significantly improved performance against 21 and 40 state-of-the-art SOD methods on eight RGB and RGB-D datasets, respectively. Consequently, we present a new SOD perspective of generating RGB-D SOD without acquiring depth data during training and testing and assist RGB methods with depth clues for improved performance. The code and trained models are available at \url{https://github.com/tanveer-hussain/EfficientSOD2}
\end{abstract}

%% file: sections/intro.tex
\section{Introduction}

\begin{figure} 
\begin{tabular}[b]{c@{}c@{}c} 
\includegraphics[width=.15\textwidth]{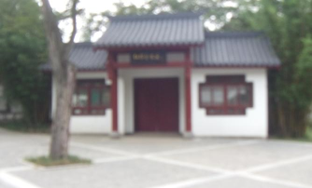}&
\includegraphics[width=.15\textwidth]{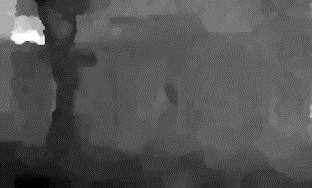}&   
\includegraphics[width=.15\textwidth]{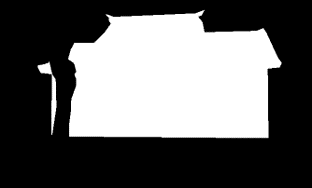} \\
Image  & Depth & GT \\ 

\includegraphics[width=.15\textwidth]{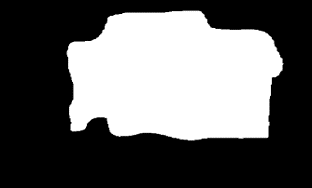}&
\includegraphics[width=.15\textwidth]{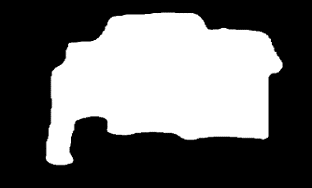}&   
\includegraphics[width=.15\textwidth]{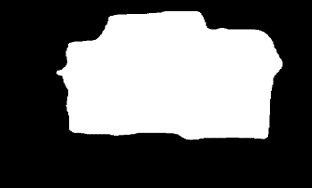} \\ 
.951  & .949 & .906 \\ 
~\ourmodel~(M1)  &~\ourmodel~(M2) &~\ourmodel~\\ 

\includegraphics[width=.15\textwidth]{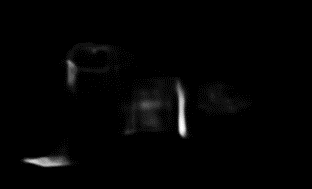}&
\includegraphics[width=.15\textwidth]{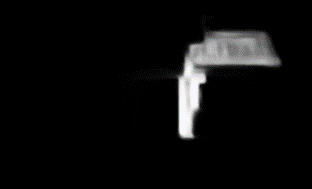}&
\includegraphics[width=.15\textwidth]{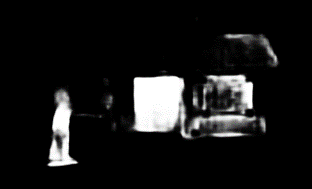} \\ 
.083  & .501 & .541 \\
BBSNet~\cite{fan2020bbs}  & PGAR~\cite{chen2020progressively} & UCNet~\cite{zhang2020uc} \\

\end{tabular}
    \caption{An example of a challenging input from RGB-D dataset with multiple candidate salient objects and noisy depth. As shown in the third row, undesirable depth has adversely affected the saliency prediction, where our ~\ourmodel~visual results in the second row are much closer to the GT with higher F-measure rates.~\ourmodel~(M1) and~\ourmodel~(M2) refer to RGB-D with original and synthetic depth.}
    \label{fig:results}
\end{figure} 

Visual saliency refers to the most noticeable and discernible contents of image data distinguishable from the background. It is often termed Salient Object Detection (SOD), as the noticeable contents are different kinds of objects inside an image. Salient objects include humans, animals, and other general object categories within an image as foreground. SOD methods~\cite{niu2012leveraging,yang2013saliency,qin2019basnet,zhu2019pdnet,fan2020bbs,wang2021salient,liu2021visual} aim to extract useful features from a given RGB or a pair of RGB and depth images to predict a binarized or gray-scale saliency map of the corresponding input. 

To achieve effective SOD, RGB or RGB-D images are processed using various soft computing techniques, where deep models~\cite{sun2021deep,cheng2014depth} have achieved the best results. Whereas, RGB-D-based SOD has recently attracted the attention of many experts due to the importance of depth data in predicting accurate saliency maps when processed composedly with RGB data, providing complementary information for objects' appearance, thereby enhancing the SOD method's performance. Moreover, the mainstream RGB-D approaches introduce multi-stream architectures to extract RGB and depth features and apply various fusion mechanisms to assist RGB data with depth clues. Previous research techniques input RGB and depth data, apply fusion at early~\cite{zhang2020uc,zhang2021uncertainty}, intermediate (multiple stages)~\cite{zhang2020asymmetric,zhang2021deep}, or later stages~\cite{wang2019adaptive} of the neural network architecture to achieve optimal detection. Recently, Zhang~\etal~\cite{zhang2020uc} integrated RGB and depth channels at the start of the network and processed six-channel input, while in their supplementary network, the authors extract distributions of RGB images as well as depth clues.

With RGB-D-based techniques, certain limitations are associated, such as additional computation is required to process depth data and, fore-mostly, the depth sensor data acquisition for generating saliency maps. Conversely, CNN-based RGB saliency models can quickly fail in challenging scenarios due to a lack of appearance cues and dependency over visual cues only. Similarly, RGB-based saliency detection methods get misled in cluttered backgrounds due to the complex salient object's appearance.

To the best of our knowledge, no single end-to-end method in SOD literature exists that takes RGB input only and generates its corresponding depth, followed by final saliency prediction using both channels. Following~\cite{xiao2018rgb}, Zhang~\etal~\cite{zhang2021deep} introduced \textit{RGB-D SOD without Depth}, a simple network that learns to predict multiple outputs during testing; but depend on the depth data for the network during training.

This article overcomes the limitations mentioned earlier by introducing novel transformer-CNN features with a pyramidal attention network for effective SOD. The proposed model, \textit{Pyramidal Attention Saliency}, abbreviated as,~\ourmodel~is specifically designed to remove the dependency over depth data, thereby increasing our model's application domains and practicability. In simple scenarios, our network's RGB flavor can be used to create accurate saliency maps. While in highly challenging scenarios, where RGB information is not sufficient for fine saliency detection, our network's RGB-D flavor generates depth from RGB and complementary features from the same depth network to support the RGB features at various levels before optimal saliency region extraction. Furthermore, it is a well-known fact that the encoder structure of the SOD model has a considerable influence while producing saliency maps as information lost during the encoding process cannot be recovered in the decoder. Therefore, along with the focus on decoder~\cite{wu2019cascaded}, the encoder part should be carefully designed, as, in~\ourmodel, we present a transformer encoder to extract multi-level features with pyramidal attention structure for image feature enhancement.

Our contributions are highlighted as follows:
\begin{itemize}
    \item We are the first to introduce the RGB-D SOD model without actually acquiring depth data as input, thus, predicting fine saliency using only single RGB input.
    \item Our encoder network considers both high and low-level CNN and transformer backbone features to process them using pyramidal transformer attention.
    \item We achieve SOTA results on both RGB and RGB-D SOD benchmarks, validating the proposed model effectiveness and practical applicability.
\end{itemize}

%% file: sections/related.tex
\section{Related Work}
The SOD techniques can be roughly categorized from the perspective of input data into i) RGB-D methods, ii) Hybrid methods, and iii) RGB methods. We provide the SOD models similar to our approach and justify the uniqueness of our model from the current state-of-the-art.

\vspace{1mm}
\noindent
\textbf{RGB-D Methods}:
The RGB-D methods need RGB and depth data for generating saliency maps, where both modalities are dependent on each other. The depth modality is specifically considered to extract the 3D layout and structural information from the input images and aid the RGB channels in generating final saliency. Currently, mainstream RGB-D SOD methods apply Convolutional Neural Networks (CNN)~\cite{piao2019depth} due to their enhanced image representation abilities, compared to handcrafted features~\cite{peng2014rgbd,feng2016local,ciptadi2013depth} which provides limited performance for complex scenarios. Some RGB-D methods merge depth and RGB channels in early stages, considering RGB-D pair as a multi-channeled  input~\cite{liu2019salient,qu2017rgbd}. Others apply late~\cite{han2017cnns} or multi-stream fusion approaches~\cite{fan2020bbs} to effectively utilize the complimentary depth information~\cite{perez2019mfas,chen2019multi,piao2019depth}. These mentioned methods require depth maps as input while generating saliency maps during training and testing. The RGB-D methods' accurate predictions are highly dependent on the depth data cues, and any noise in the depth data results in mispredictions. In some cases, particularly noisy regions are considered salient objects by the model. Thus, for an RGB-D SOD model, the dependency of depth data affects the performance in many scenarios, despite the higher computational complexity and expensive resources required for depth sensor data acquisition.

\vspace{1mm}
\noindent
\textbf{Hybrid Methods}: 
If trained on RGB-D data, the model is considered a hybrid and can predict saliency maps without acquiring depth data. These approaches tend to eliminate the dependency of supplementary depth cues while generating output. Initially, Ziao~\etal~\cite{xiao2018rgb} introduced the concept of leveraging depth information with RGB data to enhance a model's performance for SOD task, based on traditional saliency features and without any end-to-end methodology for depth and saliency prediction. Similarly, A2dele~\cite{piao2020a2dele}, an adaptive and attentive depth distiller, is introduced in recent research, aiming at the RGB-D network's efficiency and reducing the dependency over depth data during testing. Their main objective is to provide an effective network that only processes RGB stream rather than involving depth stream during the testing procedure. 

Very recently, Zhang~\etal~\cite{zhang2021deep} presented a deep RGB-D network that learns to predict saliency maps as well as depth images from an RGB-D input; thus, the network takes RGB-D data while training and does not require depth data during evaluation, but results without depth input are significantly lagging behind the RGB-D SOTA. Furthermore, deep RGB-D w$/$o depth methods still have the primary deficiency of depth sensory data dependency during training. It is not easy to install multiple sensors or a single sensor with multiple functionalities in some real-world environments due to marginally higher costs. Finally, a very recent research~\cite{zhang2021deep} fuses the actual predicted depth at the last stages, making their model biased towards the RGB channel information.

\vspace{1mm}
\noindent
\textbf{RGB Methods}:
Traditionally, salient object detection relied on the RGB input only for decades. So far, a handful CNN based research works employed RGB data only as input to produce refined saliency maps. The CNN models follow the traditional encoder-decoder architecture while generating saliency, where the network is used as an encoder to extract initial, middle, and high-level features, followed by upsampling strategies to generate saliency maps~\cite{hussain2021densely,zhao2019egnet}. Different from traditional decoders, Wu~\etal~\cite{wu2019cascaded} introduced a \textit{partial decoder} that only considers features extracted from deeper CNN layers, generating initial and final saliency maps. The initial saliency is processed applying a holistic attention module which is then fused through element-wise multiplication with encoder features and their proposed partial decoder for the final saliency map. Many researchers focus on the edge information~\cite{zhao2019egnet}, while others apply boundaries-directed features learning Siamese networks and modified fusion strategies to generate binarized saliency maps~\cite{li2021looking}.

The existing SOD literature advocates that current deep models are heavily dependent on depth clues (see Figure~\ref{fig:results} 3$^{rd}$ row) while generating refined saliency maps, mainly processing RGB and depth input via multi-stream networks. Herein, we show that providing depth data should only enhance the prediction performance (Figure~\ref{fig:results}, 2$^{nd}$ row), and refined saliency maps can be acquired using RGB input. Furthermore, RGB-D models need both RGB and depth simultaneously to predict saliency. In contrast, we propose to estimate depth from the RGB image and extract features from synthetic depth to enhance the encoder's performance. Although there are previous attempts~\cite{zhang2021deep,xiao2018rgb,piao2020a2dele} to remove the depth input reliance and ease their practical applicability, these are hybrid (RGB+RGB-D) models \ie, they process depth during training and have comparatively higher error rates and minimal performance for challenging RGB datasets.

%% file: sections/methodology.tex
\begin{figure*}
\begin{center}
\includegraphics[width=0.8\textwidth]{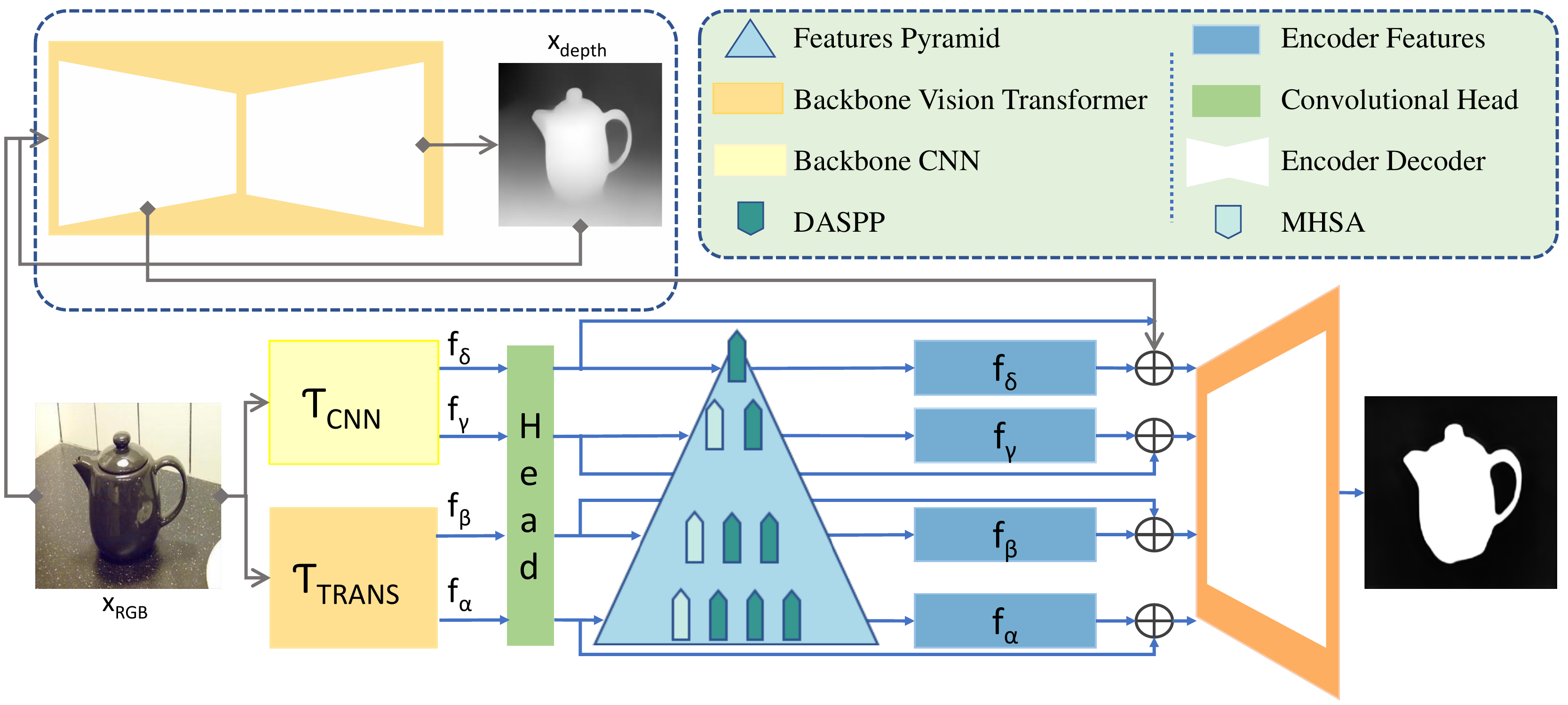}
\end{center}
\vspace{-5mm}
\captionsetup{font=small}
\caption{The proposed SOD framework. The encoder module takes an RGB image and passes it to a depth predictor ($\Delta$) to acquire $X_{depth}$, a CNN model ($\tau_{CNN}$) with a ResNet backbone produces $f_\alpha$ and $f_\beta$ feature vectors, and the transformer encoder ($\tau_{Trans}$) generates $f_\gamma$ and $f_\sigma$ features. We transform these features to equal channels using a head with sequential convolutional operations. These outputs are then transformed to highly salient features, effective for SOD using the proposed pyramidal block of DASPP and MHSA $\rho$. We employ skip connections after acquiring features from $\rho$ and finally concatenate the depth features extracted from an intermediate stage of $\Delta$ model using a synthetic depth image with the $f_\sigma$ features. The decoder module continuously fuses information, employ and upsample them gradually with residual channel attention at multiple stages before final saliency prediction. We applied a weighted fusion strategy over different losses to update our model's parameters.}
\label{fig:framework}
\end{figure*}

\section{Methodology}
In this section, we describe the proposed SOD model. We employ a pyramidal architecture to incorporate CNN and transformer backbone features. First, we comprehensively introduce our framework explain each component of our model in detail, followed by the proposed pyramid structure for feature refining and saliency extraction.

\subsection{Overview}
The proposed network consists of several components: i) feature extraction from the backbone network with customized dense connections; ii) pyramid block for encoder feature refining;  and iii) upsampling and feature fusion in the decoder. In the case of the depth requirement, our network generates a corresponding depth map using a pre-trained ViT-based monocular depth estimation model~\cite{ranftl2021vision} and extracts intermediate features from the same model to be fused with the RGB module before decoding. The overall process is visualized in Figure~\ref{fig:framework} and explained in subsequent sections.

\subsection{Pyramidal Self-Attention}
Transformer networks process bag-of-words$/$tokens representations of input data \cite{vaswani2017attention}, where image features play \textit{tokens} in case of ViT. Herein, following the findings of \cite{ranftl2021vision}, we extract deep pixel features from the input RGB image \textit{$I_{RGB_{h \times w}}$} using a pre-trained \textit{embedding network}~\cite{he2016deep} and use them as \textit{tokens} $t = \{ t_0, ... , t_{N_p}\}$, where $N_p = \frac{\textit{h} \times \textit{w}}{p^2}$, with height $h$, width $w$ and $p$ refers to the patch size considered in our network. The encoder network produces $ f_\alpha, f_\beta, f_\gamma, f_\sigma = \tau(t)$, where $\tau$ refers to the proposed encoder. The $f_\alpha, f_\beta$ are CNN features acquired from the feature embedding network's $b_i$ and $b_j$ blocks and $f_\gamma, f_\sigma$ are transformer stages $s_i$ and $s_j$. To decode $f_\gamma$ and $f_\sigma$ features from the transformer layers, we use a patch un-embedding strategy.

To enable maximum information flow between CNN features and transformer blocks, we embed dense connections as follows:

\begin{equation}
\begin{split}
    f_{\alpha} &= \tau_{CNN_{bi}} (I_{RGB}),\\
    f_{\beta}  &=\tau_{CNN_{bj}} (f_\alpha).
\end{split}
\end{equation}
So far, we have two kinds of feature vectors comprising initial edge information $(f_\alpha, f_\beta)$, we extract shapes and structure of objects, and global receptive fields $(f_\gamma, f_\sigma)$ as
\begin{equation}
\begin{split}
    f_\gamma &= \tau_{Trans_{si}} (f_\alpha\oplus f_\beta)), \\
    f_\sigma &= \tau_{Trans_{sj}} (f_\alpha\oplus f_\beta\oplus f_\gamma)),
\end{split}
\end{equation}
where $\oplus$ is the concatenation operation. These four extracted features have a common trans-head to refine and balance the channels for the proposed pyramidal attention blocks. The trans-head has three convolutional layers, each followed by batch normalization and ReLU activation.

Based on the assumption that initial low-level features $(f_\alpha, f_\beta)$ demand further refining and the later stage transformer features$(f_\gamma, f_\sigma)$  only need booster layers to enhance the existing representations. We employ pyramid attention block $P_\rho$ to finally acquire the encoder features. The $P_\rho$ contains multi-scale Dense Atrous Spatial Pyramid Pooling (DASPP) layers and multi-headed self-attention (MHSA) at the end of each $P_\rho$ block except the final $f_\sigma$ features. Since the convolutional features need additional attention, they fit into the pyramid's base with multiple DASPP modules followed by an attention mechanism. The last stages of transformer features are utilized using a single DASPP and MHSA module. As $f_\sigma$ features are already enhanced enough; therefore, only a single DASPP module is employed for further feature enhancement. The overall structure of $P_\rho$ is shown in Eq.~\ref{eq:rho}. 
\begin{equation}
\resizebox{\columnwidth}{!}{$
\left[ {\begin{array}{c}
    f_\sigma \\
    f_\gamma     \\
    f_\beta          \\
    f_\alpha

  \end{array} } \right] 
  =
  \left[ {\begin{array}{cccc}
    &       &  DASPP_{41}    &   \\
    &  DASPP_{31} & MHSA_3 &             \\
    DASPP_{21}    &   DASPP_{22}          &     MHSA_2       &             \\
    DASPP_{11}    &   DASPP_{12}          &  DASPP_{13}          &    MHSA_1 
  \end{array} } \right] $}
  \label{eq:rho}
\end{equation}

Once the refined features are extracted from the RGB input, we enrich $f_\sigma$ features via depth estimation ($\Delta$) model's features. It is noteworthy that depth is not accompanied (\ie, actual depth acquired via a depth sensor) with the input image in the RGB-D dataset; instead, it is synthetically generated from the $\Delta$ model. Due to simplicity, we do not create a separate network to process depth data individually, which can be a significant future direction for RGB-D SOD models. The concatenation at this point aids the $f_\sigma$ with fine-grained depth information, producing more refined representations (see Figure \ref{fig:results} second row).

Our main objective in the proposed encoder network is not to lose any information, as it cannot be recovered during the decoding procedure. Therefore, we consider information about edges filtered by the initial CNN blocks and employ several transformer layers to extract receptive fields to keep track of the varied size of objects. Thus, combining convolutional and transformer layers' features enables utilizing larger receptive fields when compared to regular convolutions and also design dependencies between spatially distinct features. Finally, the proposed pyramid structured attention boosts the features for the SOD problem by applying DASPP~\cite{yang2018denseaspp} for receptive fields capturing from the overall image and multi-head attention for acquiring salient information from paired feature representations.

\subsection{Fusion Attention}
Hence, the encoder network is very dense, extracting various features. Therefore, we only focus on fusing the encoder features in the decoder network before generating the final saliency. The decoder receives $f_\alpha, f_\beta, f_\gamma$, and $f_\sigma$ features, progressively concatenating from the fusion of shallower to deeper layers. After each concatenation branch, our network employs the \emph{residual channel attention} module to capture spatial and channel dependencies between pixel locations and different channels of the input feature maps. We achieve the final saliency map $S$ by gradually upsampling the features during concatenation, as given below mathematically.

\begin{equation}
\begin{split}
   S_{i}    &=   \theta (f_\sigma \oplus f_\gamma),        \\
   S_{i+1} &=    \theta (\cup (S_{i} \oplus f_\beta),       \\
   S_{i+2} &=    \theta (S_{i+1} \oplus f_\alpha),     \\
   S       &= \cup_{h \times w} (C(S_{i+2})),
 \end{split}
\end{equation}
where $\cup$ is the upsampling operation, $\theta$ refers to the residual channel attention module, and $C$ is the balancing convolution to generate a single channel saliency map $S$.

\begin{figure*} [t]
\centering
\begin{tabular}[b]{c@{}c@{}c@{}c@{}c@{}c@{}c@{}c@{}c@{}c} 
\raisebox{0.1\normalbaselineskip}[0pt][0pt]{\rotatebox{90}{DUT-RGBD}}&  \hspace{0.02mm}    
\includegraphics[width=.1\textwidth]{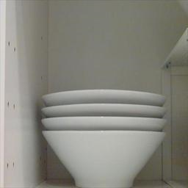} \hspace{1mm}&
\includegraphics[width=.1\textwidth]{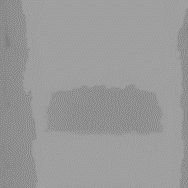} \hspace{1mm} &   
\includegraphics[width=.1\textwidth]{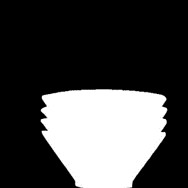} \hspace{1mm} &
\includegraphics[width=.1\textwidth]{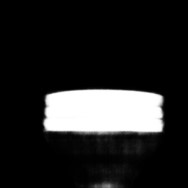} \hspace{1mm} &
\includegraphics[width=.1\textwidth]{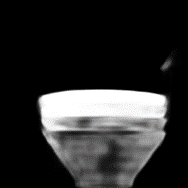} \hspace{1mm} &
\includegraphics[width=.1\textwidth]{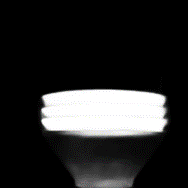} \hspace{1mm} &   
\includegraphics[width=.1\textwidth]{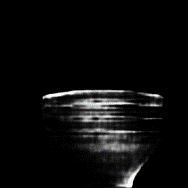} \hspace{1mm} &
\includegraphics[width=.1\textwidth]{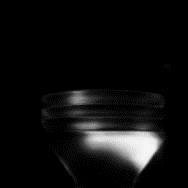} \hspace{1mm}&
\includegraphics[width=.1\textwidth]{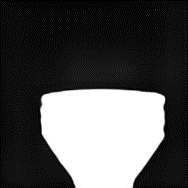}\\

\raisebox{0.2\normalbaselineskip}[0pt][0pt]{\rotatebox{90}{NJUD2K}}& \hspace{0.02mm}
\includegraphics[width=.1\textwidth]{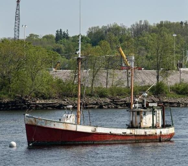} \hspace{1mm}&
\includegraphics[width=.1\textwidth]{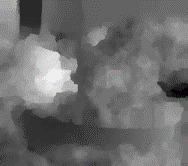} \hspace{1mm} &   
\includegraphics[width=.1\textwidth]{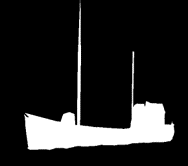} \hspace{1mm} &
\includegraphics[width=.1\textwidth]{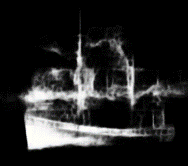} \hspace{1mm} &
\includegraphics[width=.1\textwidth]{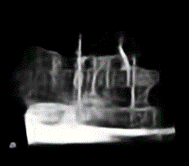} \hspace{1mm} &
\includegraphics[width=.1\textwidth]{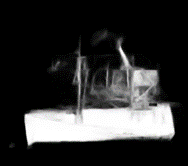} \hspace{1mm} &   
\includegraphics[width=.1\textwidth]{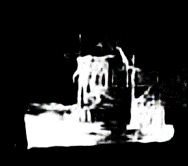} \hspace{1mm} &
\includegraphics[width=.1\textwidth]{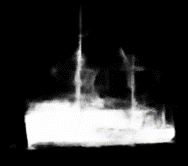} \hspace{1mm}&
\includegraphics[width=.1\textwidth]{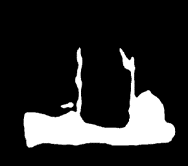}\\

\raisebox{1.5\normalbaselineskip}[0pt][0pt]{\rotatebox{90}{NLPR}}& \hspace{0.02mm}
\includegraphics[width=.1\textwidth,height=.11\textwidth]{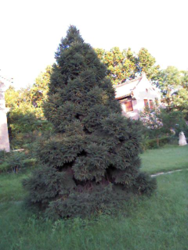} \hspace{1mm}&
\includegraphics[width=.1\textwidth,height=.11\textwidth]{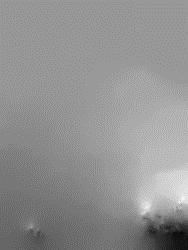} \hspace{1mm} &   
\includegraphics[width=.1\textwidth,height=.11\textwidth]{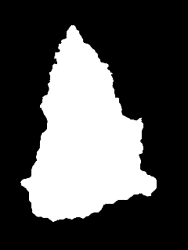} \hspace{1mm} &
\includegraphics[width=.1\textwidth,height=.11\textwidth]{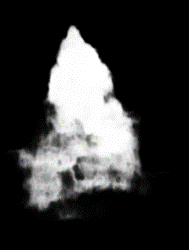} \hspace{1mm} &
\includegraphics[width=.1\textwidth,height=.11\textwidth]{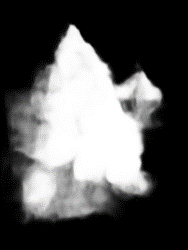} \hspace{1mm} &
\includegraphics[width=.1\textwidth,height=.11\textwidth]{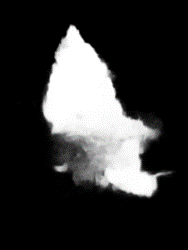} \hspace{1mm} &   
\includegraphics[width=.1\textwidth,height=.11\textwidth]{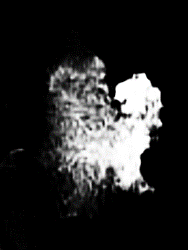} \hspace{1mm} &
\includegraphics[width=.1\textwidth,height=.11\textwidth]{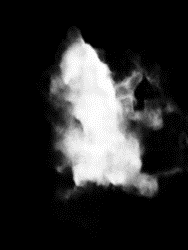} \hspace{1mm}&
\includegraphics[width=.1\textwidth,height=.11\textwidth]{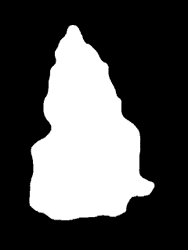}\\

\raisebox{2\normalbaselineskip}[0pt][0pt]{\rotatebox{90}{SIP}}& \hspace{0.02mm}
\includegraphics[width=.1\textwidth,height=.11\textwidth]{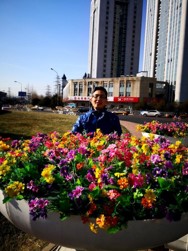} \hspace{1mm}&
\includegraphics[width=.1\textwidth,height=.11\textwidth]{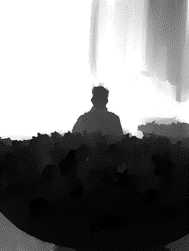} \hspace{1mm} &   
\includegraphics[width=.1\textwidth,height=.11\textwidth]{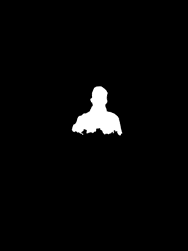} \hspace{1mm} &
\includegraphics[width=.1\textwidth,height=.11\textwidth]{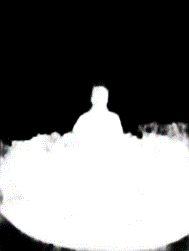} \hspace{1mm} &
\includegraphics[width=.1\textwidth,height=.11\textwidth]{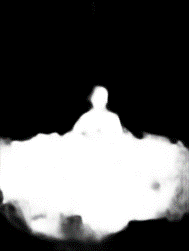} \hspace{1mm} &
\includegraphics[width=.1\textwidth,height=.11\textwidth]{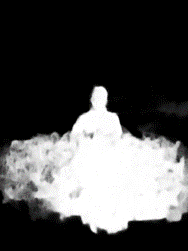} \hspace{1mm} &   
\includegraphics[width=.1\textwidth,height=.11\textwidth]{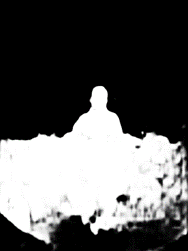} \hspace{1mm} &
\includegraphics[width=.1\textwidth,height=.11\textwidth]{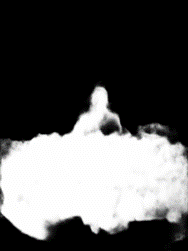} \hspace{1mm}&
\includegraphics[width=.1\textwidth,height=.11\textwidth]{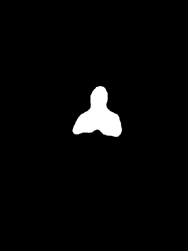}\\

& Image  & Depth & GT & DCF~\cite{ji2021calibrated} & BBSNet~\cite{fan2020bbs} & PGAR~\cite{chen2020progressively} & UCNet~\cite{zhang2020uc} & D3Net~\cite{fan2020rethinking} &~\ourmodel~\\
\end{tabular}
    \caption{Visual comparison of our~\ourmodel~with SOTA models using the most challenging images from RGB-D datasets. Our results are more close to the groundtruth. Our saliency maps have fine details and clear edges as compared to the state-of-the-art algorithms.}
    \label{fig:visualcomparison}
    \vspace{-3mm}
\end{figure*}

\subsection{Objective Function}
The proposed objective function is the weighted fusion of several loss functions defined below for input $y$ and ground truth $x$ images. 
\begin{equation} 
\ell_{{total}_{(y,x)}} = \epsilon_1 \ell_{st} + \epsilon_2 \ell_{ssim} + \epsilon_3 \ell_{2}+ \epsilon_4 \ell_{se} ,
\label{eq:TotalLoss} 
\end{equation}
where $\epsilon$ is the constant having values of $\epsilon_1=0.2$, $\epsilon_2=0.3$, $\epsilon_3=0.2$, and $\epsilon_4=0.3$. 

The structure loss $\ell_{st}$ focuses on global structure optimization instead of aiming at a single pixel, thereby being significantly immune towards unbalanced distributions. We adopt structure loss from~\cite{wei2020f3net}, where weights are assigned to hard pixels to highlight their importance, instead of treating all pixels equally and ignoring the difference between pixels. The $\ell_{st}$ loss is given as 
\begin{equation} 
\ell_{{st}_{(y,x)}} = 1 - \frac{\sum_{i=1}^{H}\sum_{j=1}^{W}(x_{ij}y_{ij})(1+\Psi\omega_{ij})}
{\sum_{i=1}^{H}\sum_{j=1}^{W}(x_{ij}+y_{ij}-x_{ij}y_{ij})(1+\Psi\omega_{ij})},
\label{eq:STRUCTLoss} 
\end{equation} 
where $\Psi$ is a hyperparameter and $\omega$ are the weights.

Furthermore, different from pixel-wise comparison losses such as the Euclidean distance, we employ Structural Similarity Index Measure (SSIM) as a loss function to compute the negation of similarity between $y$ and $x$ and update the model based on their difference. 

\begin{equation} 
\ell_{{ssim}_{(y,x)}} = 1 - \frac{(2\mu_y\mu_{x}+c_1)(2\varsigma_y{yx}+c_2)} {(\mu_y^2 + \mu_{x}^2 + c_1)(\varsigma_y^2+\varsigma_y{x}^2+c_2)}.
\label{eq:SSIMLoss} 
\end{equation} 

In Eq.~\ref{eq:SSIMLoss}, $\mu_i$ and $\mu_t$ represent the average, $\varsigma_{it}$ is the covariance, $\varsigma_i^2$ and $\varsigma_t^2$ represent variance of input $y$ and the ground truth image $x$, respectively. Herein, $c_1 = (k_1,L)^2$, $c_2 = (k_2,L)^2$ that are two variables to neutralize the division in case of weak denominator. Finally, $L$ is a dynamic range of pixel values and $k_1=0.01, k_2=0.03$. 

Moreover, $\ell_{2}$ refers to $\ell_{ssim}$ regularization. Due to the involvement of multiple features, our network becomes complex and prone to overfitting; therefore, we adopt $\ell_2$ regularization in our loss function to avoid such problems.
Finally, edge-aware smoothness loss is adopted from~\cite{godard2017unsupervised,wang2018occlusion}, referred to as $\ell_{se}$ in Eq.~\ref{eq:TotalLoss}, which makes the distinctions between objects smoother using disparity gradients. Herein, $\epsilon$ is used to assign weights multiplied with each loss value when calculating the final objective function for the model update.

\subsection{Implementation Details} 
Inspired by the CNN-based SOD models'~\cite{zhang2021uncertainty,fan2020rethinking,sun2021deep} backbone strategies, we employ pre-trained DPT~\cite{ranftl2021vision}, that is trained on a large-scale semantic segmentation dataset as the ViT networks perform well when trained using large-scale datasets~\cite{dosovitskiy2020image}. For feature extraction, following~\cite{ranftl2021vision}, we extract transformer tokens from a ResNet50 pre-trained model~\cite{he2016deep}. We utilize the same model's initial two blocks $b_i, b_j=1,2$ to extract $f_\alpha, f_\beta$ features. In case $\tau_{TRANS}$, $f_\gamma, f_\sigma$ are transformer stages $s_i, s_j = 9, 12$ features. It should be noted that we embed depth transformer features from stage $s=12$ of depth estimator backbone model $\Delta$ into $f_\sigma$ features to enhance the representations of existing features. The resolution of output features is $h\times\frac{1}{2^n}$, where $n=1,2,3,$ and $ 4$ for $f_\alpha, f_\beta, f_\gamma,$ and $f_\sigma$, respectively with $D_{feat}=256$ features' dimensions.

%% file: sections/experiments.tex
\input{sections/tables}

\section{Experiments}

\subsection{Setup}

\noindent
\textbf{Training details:}
We implement our method in the PyTorch deep learning framework with an NVIDIA GeForce RTX 3090 GPU. Our initial training settings are inspired from~\cite{zhang2020uc}. We use an ADAM optimizer with 0.9 momentum, the initial learning rate is set to 5$e^{-5}$, and the model is trained for 25 epochs. The training images are resized to a standard 224 $\times$ 224. The ResNet50~\cite{he2016deep} is used as the backbone with the pre-trained ImageNet weights in the CNN module, while in the Transformer module, we use~\cite{ranftl2021vision}. After each epoch, the learning rate is adjusted with a 10\% decrease, and the batch size is six.

\vspace{1mm}
\noindent
\textbf{Datasets:} We carry out experiments on four benchmark RGB-D datasets including three standard such as DUT-RGBD \cite{piao2019depth}, NJUD2K~\cite{ju2014depth}, NLPR~\cite{peng2014rgbd}, and newly introduced SIP~\cite{fan2020rethinking}. We also compare on four datasets, that are widely used for RGB SOD method evaluation, including ECSSD~\cite{yan2013hierarchical}, HKU-IS~\cite{li2015visual}, DUTS-TE~\cite{wang2017learning}, and PASCAL-S~\cite{li2014secrets}. We follow the provided training and testing dataset split for each one~\cite{fan2020rethinking}.

\vspace{1mm}
\noindent
\textbf{Evaluation Metrics}:
We evaluate our methods on three widely used metrics: Mean Absolute Error (\textit{MAE}), standard F-measure (\textit{F$_\beta$}), and E-measure (\textit{E$_\alpha$}). We have verified the results of our methods via MATLAB\footnote{https://github.com/jiwei0921/Saliency-Evaluation-Toolbox/} and Python\footnote{https://github.com/taozh2017/SPNet/blob/main/Code/utils/evaluator.py} evaluation toolboxes. It should be noted that the lower the (\textit{MAE}) the better it is and vice versa for (\textit{F$_\beta$})  (\textit{S$_\alpha$}), and (\textit{E$_\alpha$}).

\subsection{Comparisons}
We compare our results with \emph{41 methods} on RGB-D datasets and \emph{21} algorithms employing RGB datasets, comprising recent deep learning-based models.

\begin{figure} 
\centering
\begin{tabular}[b]{c@{ }c@{ }c@{ }c@{ }c@{ }c} 
\includegraphics[width=.16\columnwidth,height=.16\columnwidth]{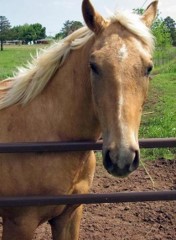}&
\includegraphics[width=.16\columnwidth,height=.16\columnwidth]{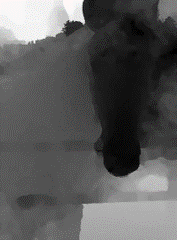}&   
\includegraphics[width=.16\columnwidth,height=.16\columnwidth]{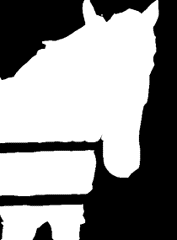}&   
\includegraphics[width=.16\columnwidth,height=.16\columnwidth]{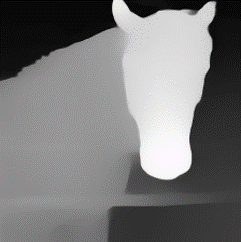}&   
\includegraphics[width=.16\columnwidth,height=.16\columnwidth]{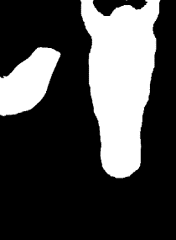} &   
\includegraphics[width=.16\columnwidth,height=.16\columnwidth]{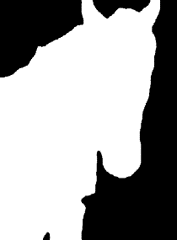}\\

\includegraphics[width=.16\columnwidth]{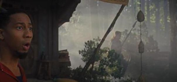}&
\includegraphics[width=.16\columnwidth]{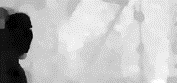}&   
\includegraphics[width=.16\columnwidth]{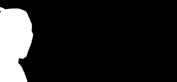}&   
\includegraphics[width=.16\columnwidth,height=.035\textwidth]{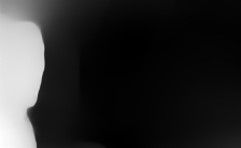}&   
\includegraphics[width=.16\columnwidth]{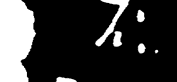} &   
\includegraphics[width=.16\columnwidth]{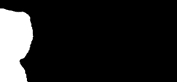}\\
Image & Depth & GT & SD & M2 & M1 \\

\end{tabular}
    \caption{Visual scenarios with $M2$'s more accurate predictions when compared to $M1$. The clearly observable difference is due to the noise in input depth data. SD refers to synthetic depth.}
    \label{fig:m1vsm2}
\end{figure}

\subsubsection{Quantitative Comparisons}
\vspace{1mm}
\noindent
\textbf{RGB-D Saliency:}
For RGB-D SOD, we compare our method with \emph{41 deep learning models}, and since traditional hand-crafted feature-based methods are not effective and accurate enough, we exclude them for simplicity. To ensure a fair comparison, we mostly report the results of these methods from their papers or follow existing articles~\cite{ji2021calibrated,zhang2021uncertainty,liu2021visual}. Furthermore, it is worth noting that there are ten RGB methods performing experiments on RGB-D datasets, while the closest works to our approach are considered hybrid models.

The performance comparisons on the RGB-D datasets are given in Table~\ref{tab:RGB-D}, where we reported three sections due to the difference in the SOD input. When compared to RGB and hybrid methods utilizing RGB-D datasets, we achieved the best results on all reported metrics,  surpassing SOTA methods by a large margin on some metrics. Similarly, we receive better performance when we generate saliency using RGB-D with original depth~\ie second best results on DUT-RGBD~\cite{piao2019depth}. On the NJUD~\cite{ju2014depth} dataset, we lag behind some current methods, but it is noteworthy that we did not use any multi-stream architecture to process depth data, as our primary objective is to eradicate dependency over depth data. On the NLPR~\cite{peng2014rgbd} dataset, we achieved SOTA on MAE and $E_\alpha$ and held second best in terms of F-measure, as~\cite{zhou2021specificity} achieved 0.004 higher $F_\beta$ from our model. Finally, we set a new SOTA over SIP~\cite{fan2020rethinking} dataset.

\vspace{1mm}
\noindent
\textbf{RGB Saliency:}
Our proposed model is compared with \emph{21 RGB-based SOD methods} in Table \ref{tab:RGB} on RGB datasets, which shows significantly reduced error rates and top performance on three ECSSD \cite{yan2013hierarchical}, HKU-IS~\cite{li2015visual}, and PASCAL-S~\cite{li2014secrets} datasets except DUTS-TE~\cite{wang2017learning}. Where~\ourmodel~is ranked as the third-best against competing methods. Thus, it can be concluded that the proposed model is also applicable in challenging scenarios that can be effectively functional and independent of the depth module.

\subsubsection{Qualitative Comparisons}
We present visual outcomes of our model against SOTA for four challenging images with noisy depth and complex background, similar or multiple salient objects,  and distant objects. Figure~\ref{fig:visualcomparison} shows how accurately~\ourmodel~predicts saliency maps.~\ourmodel~ predicts high-quality, smooth saliency of the input image with similar background and foreground (1$^{st}$ and 3$^{rd}$ rows), where top-ranked RGB-D models produced very coarse predictions. Likewise,~\ourmodel~ correctly identified distant objects (2$^{nd}$ row) well, even though the accompanying depth of corresponding RGB has much noise, confusing the depth-dependent deep models to produce non-smooth dispersed predictions. Similarly, for an object hiding behind informative background (4$^{th}$ row), our method produced exactly similar saliency as ground truth, whereas the compared SOTA produced limited results. It should be noted that our method produces similar or very close saliency maps to the ground truth. Therefore, analyzing qualitative and quantitative results demonstrates the proposed model's effectiveness and robustness.

\subsubsection{Ablation Study}
\begin{figure}[hbt!]
    \centering
    \begin{tikzpicture}
        \begin{axis} 
        [ybar,
        width=\columnwidth,
        height=200,
        ymajorgrids=true,
        xmajorgrids=false,
        xminorgrids=true,
        yminorgrids=false,
        minor tick num=1,
        ylabel={\ MAE},
        xlabel={\ Methods},
        yticklabel style={%
                 /pgf/number format/.cd,
                     fixed,
                     fixed zerofill,
                     precision=2,
                     },
        enlarge x limits = 0.15,
        bar width = 8pt,
        xtick=data,
        ymax=.1,
        legend style={at={(0.32,0.99)},anchor=north},
        symbolic x coords=
        {
        M1, %
        M2,
        M3,
        PASNet,
        },
        nodes near coords,
        every node near coord/.append style={rotate=90, anchor=west,/pgf/number format/fixed zerofill,
        /pgf/number format/precision=4}
        ]
        
        \addplot 
        coordinates 
        {
        (M1,.028)
        (M2,.029)
        (M3,.074)
        (PASNet,.033)
        }; 
        
        \addplot 
        coordinates 
        {
        (M1,.051)
        (M2,.040)
        (M3,.067)
        (PASNet,.040)
        };  
        
        \addplot 
        coordinates 
        {
        (M1,.021)
        (M2,.021)
        (M3,.073)
        (PASNet,.024)
        }; 
        
        \addplot 
        coordinates 
        {
        (M1,.016)
        (M2,.015)
        (M3,.071)
        (PASNet,.018)
        };

        \legend  {DUT-RGBD,NJUD,NLPR,SIP}
        
    \end{axis}
    \end{tikzpicture}

    \caption{Error rates of ablation studies conducted using RGB-D datasets.}
    \label{fig:ablation}
\end{figure}
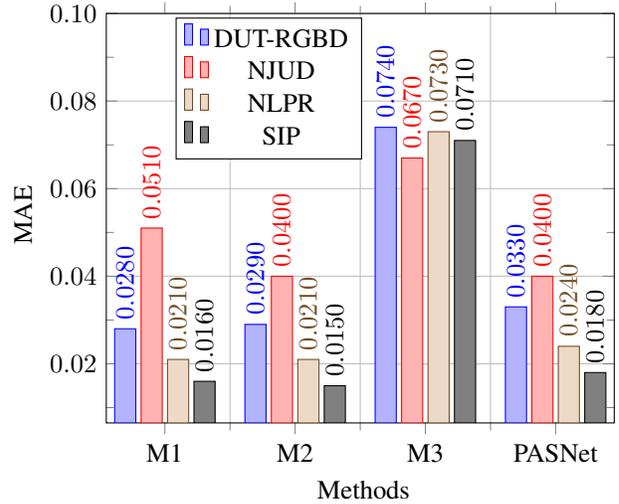

We carried out several ablation studies to analyze the effect of the various parameters for the proposed framework. $M1$ and $M2$ refer to experimentation in terms of data input \ie, including depth data features to predict the final saliency, provided in Table~\ref{tab:RGB-D}. $M3$ refers to the effect of the proposed pyramidal attention block, where visual qualitative results are presented in Figure~\ref{fig:m1vsm2}, and detailed quantitative ablation results of various models are given in Figure~\ref{fig:ablation}. Next, we provide our analysis.

\vspace{1mm}
\noindent
\textbf{Original vs. Synthetic Depth:}
Basically, the proposed model processes only RGB input, where in some cases, we integrated depth data to see the effect on output predictions. Depth data for sure increases the smoothness of saliency prediction, and hence it improves the performance (see Figure~\ref{fig:m1vsm2}). Firstly for $M1$, we used the original depth data provided in RGB-D datasets and achieved excellent performance compared to SOTA and other variants of our model, but it comes at the cost of depth data dependency. Next, for $M2$, we generated depth from the input RGB image and achieved a comparable decrease in performance against $M1$, but still, this model outperformed SOTA models of its kind. In the case of NJUD2K and SIP datasets, $M2$ has better performance due to noise in the original depth, as shown in Figure~\ref{fig:m1vsm2}. Although our RGB-D methods have a lower dependency on depth features, the saliency prediction performance is still adversely affected by noise or improper depth information.

\vspace{1mm}
\noindent
\textbf{Effect of Proposed Pyramidal Attention ($P_\rho$):}
To analyze the impact of the proposed pyramidal attention block features performance, we exclude $P_\rho$ from~\ourmodel~named as $M3$. It is observable from Figure~\ref{fig:ablation} that the error rates significantly increase due to the non-effective feature representations directly acquired from ViT and CNN networks without post-enhancement using $P_\rho$. The pyramidal attention block reduces the error rates by an average of 4.525\% on four RGB-D datasets.


%% file: sections/tables.tex
\begin{table*}[t]
\caption{Quantitative results of our algorithm against competing methods. The symbols E$_\alpha$, F$_\beta$, and S$_\gamma$ refer to the E-measure, F-measure, and S-measure. The (\textcolor{red}{Red} color represents the best results and the \textcolor{blue}{Blue} color is for the second-best results in each block \ie RGB-D, RGB, and Hybrid models.}
\captionsetup{justification=centering}
\begin{center}
\resizebox{\textwidth}{!}{
\begin{tabular}[width=0.9\columnwidth]{cl|llll|cccc|cccc|cccc}

\toprule

& \multirow{2}{*}{Methods} & \multicolumn{4}{c|}{DUT-RGBD \cite{piao2019depth}} & \multicolumn{4}{c|}{NJUD2K \cite{ju2014depth}} & \multicolumn{4}{c|}{NLPR \cite{peng2014rgbd}} & \multicolumn{4}{c}{SIP \cite{fan2020rethinking}} \\

& & E$_\alpha\uparrow$    &   F$_\beta\uparrow$    & S$_\gamma\uparrow$ & MAE$\downarrow$    &   E$_\alpha\uparrow$    &   F$_\beta\uparrow$    & S$_\gamma\uparrow$ &  MAE$\downarrow$    &      E$_\alpha \uparrow$    &   F$_\beta\uparrow$    & S$_\gamma\uparrow$ &  MAE$\downarrow$    &     E$_\alpha \uparrow$    &   F$_\beta\uparrow$    & S$_\gamma\uparrow$ &  MAE$\downarrow$  \\ \midrule
 

\multirow{30}{*}{\rotatebox{90}{RGB-D}} & DF \cite{qu2017rgbd}  &  .     &  .465     & . &   .185   &  .700     &  .653     & .763 &  .140    &  .757     &   .664    & .806 &  .079    &  .565     &  .465     & .653 &  .185   \\

& CTMF \cite{han2017cnns} &  .     &  .608     & . &   .139   &  .846     &  .779     & .849 &  .085    &  .840     &   .740    & .860 &  .056    &  .704     &  .608     & .716 &  .139  \\ 

& PCF \cite{chen2018progressively} &  .     &  .814     & . &   .071   &  .895     &  .840     & .877 &  .059    &  .887     &   .802    & .874 &  .044    &  .878     &  .814     & .842 &  .071  \\ 

& MMCI \cite{chen2019multi} &  .     &  .771     & . &   .086   &  .851     &  .793     & .858 &  .079    &  .841     &   .737    & .856 &  .059    &  .845     &  .771    & .833  &  .086   \\

& CPFP \cite{zhao2019contrast}  &  .     &  .821     & . &   .064   &  .910     &  .850     & .878 &  .053    &  .918     &   .840    & .888 &  .053    &  .893     &  .821     & .850 &  .064   \\

& CPD \cite{wu2019cascaded} &  .     &  .872     & . & .042   &  .     &  .874     & . &  .051    &  .     &   .878   & .  &  .028    &  .     &  .884     & . &  .043  \\ 

& TANet \cite{chen2019three} &  .     &  .803     & . &   .075   &  .895     &  .841     & .879 &  .059    &  .902     &   .802    & .886 &  .044    &  .870     &  .803     & .835 &  .075  \\

& AFNet \cite{wang2019adaptive} &  .     &  .702     & . &   .118   &  .867     &  .827     & .822 &  .077    &  .851     &   .755    & .799 &  .058    &  .793     &  .702     & .720 &  .118  \\ 

& DMRA \cite{piao2019depth} &  . 888    &  .883     & \textcolor{blue}{.927} &   .048   &  .920     &  .873     & .886 &  .051    &  .940     &   .865   & .899  &  .031     &  .844     &  .811    & .806  &  .085  \\

& ATSA  \cite{zhang2020asymmetric} &  .  &  .918     & .916 &  .032   &  .     &  893    & .885 &  .040    &  .     &   .876    & .909 &  .028    &  .     &  .871     & .849 &  .053  \\

& PGAR  \cite{chen2020progressively} &  .933  &  .914     & .899 &  .035   &  .940     &  .893    & .909 &  .042    &  .     &   .885    & .917 &  .024    &  .     &  .854     & .838 &  .055  \\

& UCNet \cite{zhang2020uc} &  .     &  .     & .871 &   .   &  .936     &  .886    & .897 &  .043    &  .951     &   .831    & .920 &  .025     &  .914     &  .867     & .875 &  .051   \\

& S$^2$MA \cite{liu2020learning}  &  .935     &  .899     & .904 &   .043   &  .930     &  .889    & .894 &  .054    &  .953     &   .902    & .916 &  .030     &  .919     &  .877     & .872 &  .058   \\

& JL-DCF \cite{fu2020jl} &  .941     &  .910     & .906 &   .042   &  .944     &  .904    & .902 &  .041    &  \textcolor{blue}{.963}     &   .918    & .925 &  \textcolor{blue}{.022}     &  .919     &  .877     & .880 &  .058   \\

& SSF-RGBD \cite{zhang2020select}  &  .950     &  .923     & .915 &   .033   &  .935     &  .896    & .899 &  .043    &  .953     &   .896    & .915 &  .027     &  .870     &  .786     & .799 &  .091   \\

& BBSNet \cite{fan2020bbs}  &  .912    &  .870     & .923 &   .058   &  \textcolor{black}{.949}     &  .919    & .921 &  \textcolor{blue}{.035}    &  .961     &   .918    & .931 &  .023     &  .922     &  .884     & .879 &  .055   \\

& Cas-Gnn \cite{luo2020cascade} &  .953    &  .926     & .920 &   .030   &  .948     &  .911    & .911 &  .039    &  \textcolor{black}{.955}     &   .906    & .923 &  \textcolor{black}{.025}  &  .919     &  .879     & .875 &  .051   \\

& CMW \cite{li2020cross} &  .864    &  .779     & .797 &   .098   &  .927     &  .871    & .870 &  .061    &  .951     &   .903    & .917 &  .029  &  .804     &  .677     & .705 &  .141   \\

& DANet \cite{zhao2020single} &  .939    &  .904     & .899 &   .042   &  .935     &  .898    & .899 &  \textcolor{black}{.046}    &  .955     &   .904    & .920 &  .028  &  .918     &  .876     & .875 &  .055   \\

& cmMS \cite{li2020rgb} &  .940    &  .913     & .912 &   .036   &  .897     &  .936    & .900 &  .044    &  .955     &   .904    & .919 &  .028  &  .911     &  .876     & .872 &  .058   \\

& UCNet-2 \cite{zhang2021uncertainty}  &  .  &  .864     & . &   .034   &  .937     &  .893     & .902 &  .039    &  .952     &   .893   & .917  &  .025    &  .927     &  .877    & .883  &  .045  \\

& SP-Net \cite{zhou2021specificity}   &  .  &  .     & . &  .   &  \textcolor{red}{.954}     &  \textcolor{red}{.935}    & \textcolor{red}{.925}  &  \textcolor{red}{.028}    &  \textcolor{black}{.959}     &   \textcolor{red}{.925}    & \textcolor{blue}{.927} &  \textcolor{red}{.021}    &  \textcolor{black}{.930}     &  \textcolor{blue}{.916}    & .894  &  .043  \\ 

& DA-MMFF  \cite{sun2021deep}  &  .950  &  .926    & .921  &  .030   &  .923     &  .901     & .903 &  .039    &  .950     &   .897    & .918 &  .024    &  .     &  .    & .  &  .  \\ 

& DCF  \cite{ji2021calibrated}  &  .952  &  .926    & .  &  .030   &  .922     &  .897     & . &  .038    &  .956     &   .893   & .  &  .023    &  .920     &  .877    & .  &  .051  \\ 

& HFNet  \cite{zhou2021hfnet}  &  .934  &  .885     & .900 & .044 &  .902     &  .859    & .898  &  .053    &  .934     &   .839   & .897  &  .038    &  .904     &  .850    & .857  &  .071  \\ 

& VST  \cite{liu2021visual}   & \textcolor{red}{.969} &  \textcolor{red}{.948} &  \textcolor{red}{.943}     &  \textcolor{red}{.024}    & %
\textcolor{blue}{.951}  &  \textcolor{blue}{.920}    &  \textcolor{blue}{.922}     &   \textcolor{blue}{.035} %
& \textcolor{black}{.962}  &  .920    &  \textcolor{red}{.943}     &  .024   & 
\textcolor{blue}{.944} & .915 & \textcolor{blue}{.904}  &  \textcolor{blue}{.040}  \\ 
\cmidrule{2-18}

& \textbf{\ourmodel$_{M1}$} &   \textcolor{blue}{.966}    &  \textcolor{blue}{.944}    & .917  &  \textcolor{blue}{.028}    &
\textcolor{black}{.938}   &   .892 & .867   &   .051   & 
\textcolor{red}{.966}    &   \textcolor{blue}{.921}  & .913   &   \textcolor{red}{.021}   &   
\textcolor{red}{.987}   &   \textcolor{red}{.956}  & \textcolor{red}{.936}   & \textcolor{red}{.016} \\

\hline


\multirow{11}{*}{\rotatebox{90}{RGB}} & DSS \cite{hou2017deeply} &  .     &  .732     & .803 &   .127   &  .     &  .776    & .769  &  .108    &  .     &   .755   & .838  &  .076    &  .     &  .   & .   &  .   \\

& Amulet \cite{zhang2017amulet}  .  &  .     &  .803   & .813   &   .083   &  .     &  .798     & .827 &  .085    &  .     &   .722   & .838  &  .062    &  .     &  .   & .   &  .  \\ 

& R$^3$Net \cite{deng2018r3net}  .  &  .     &  .781    & .819  &   .113   &  .     &  .775    & .770  &  .092    &  .     &   .649   & .846  &  .101    &  .     &  .   & .   &  .  \\ 

& PiCANet \cite{liu2018picanet} .  &  .     &  .826    & .878  &   .080   &  .     &  .806    & .872  &  .071    &  .     &   .761  & .871   &  .053    &  .     &  .   & .   &  .  \\ 

& PAGRN \cite{zhao2019pyramid}  .  &  .     &  .836    & .  &   .079   &  .     &  .827   & .   &  .081    &  .     &   .795   & .  &  .051    &  .     &  .    & .  &  .  \\ 

& PoolNet \cite{liu2019simple}  .  &  .     &  .871    & .  &   .049   &  .     &  .850    & . &  .057    &  .     &   .791   & .  &  .046    &  .     &  .   & .   &  .  \\ 

& AFNet \cite{feng2019attentive}  .  &  .     &  .851    & .  &   .064   &  .     &  .857   & .  &  .056    &  .     &   .807   & .  &  .043    &  .     &  .    & .  &  .  \\ 

& CPD \cite{wu2019cascaded}  . &  .911  &  .865    & .875  &  .055 &  .905     &  .905    & .863  &  .060    &  .925     &   .866   & .893  &  .034    &  .     &  .    & .  &  .  \\

& EGNet \cite{zhao2019egnet}  .  &  .     &  .866    & .872  &   .059   &  .     &  .846    & .840  &  .060    &  .     &   .800   & .880  &  .047    &  .     &  .876   & .   &  .049  \\

& MSI-Net \cite{pang2020multi}  .  &  .900     &  .861    & .875  &   .058   &  .906     &  .859    & .870  &  .057    &  .914     &   .854   & .886  &  .041    &  .     &  .   & .   &  .  \\

& DCF  \cite{ji2021calibrated}  . &  .  &  .    & .  &  .   &  .     &  .869    & .  &  .046    &  .     &   .855  & .   &  .028    &  .     &  .839    & .  &  .063  \\ 

\cmidrule{2-18}

& \textbf{\ourmodel} &   \textcolor{red}{.966}     &  \textcolor{red}{.940}    & \textcolor{red}{.903}  &   \textcolor{red}{.033}   &  \textcolor{red}{.946}     &  \textcolor{red}{.907}     & \textcolor{red}{.891} &  \textcolor{red}{.040}    &  \textcolor{red}{.964}     &   \textcolor{red}{.921}   & \textcolor{red}{.912}  &  \textcolor{red}{.024}    &  \textcolor{red}{.987}     &  \textcolor{red}{.955}    & \textcolor{red}{.930}  &  \textcolor{red}{.018}  \\

\hline

\multirow{5}{*}{\rotatebox{90}{Hybrid}} & A2dele \cite{piao2020a2dele} &  .     &  \textcolor{blue}{.892}     & \textcolor{blue}{.885} &   \textcolor{blue}{.042}   &  .     &  .884    & .871  &  .051    &  .     &   .878  & .899   &  \textcolor{blue}{.028}    &  .     &  .884   & .829   &  .043  \\

& DASNet \cite{zhao2020depth} &  .     &  .    & .  &   .   &  .     &  .894   & \textcolor{red}{.902}   &  \textcolor{blue}{.042}    &  .     &   \textcolor{blue}{.907}   & .929  &  \textcolor{red}{.021}    &  .     &  .    & .  &  .  \\

& DeepRGB-D \cite{zhang2021deep} & \textcolor{blue}{.902} & .853  &  .864     &  .072    & .927  &   .876   &  .886     &  .050    & .936  &  .882    &  .906     &   .038   & .  &  .    &  .     &  .     \\

 \cmidrule{2-18}

& \textbf{\ourmodel$_{M2}$}  & \textcolor{red}{.966}    &  \textcolor{red}{.942}    & \textcolor{red}{.903}  &  \textcolor{red}{.029}    &  \textcolor{red}{.948}     &   \textcolor{red}{.908}   & \textcolor{blue}{.891}  &   \textcolor{red}{.040 }  &   \textcolor{red}{.962}    &   \textcolor{red}{.917}   & \textcolor{red}{.912}  &   \textcolor{red}{.021}   &    \textcolor{red}{.988}   &   \textcolor{red}{.958}   & \textcolor{red}{.930}  & \textcolor{red}{.015} \\ \bottomrule

\end{tabular}
}
\end{center}
\label{tab:RGB-D}
\end{table*}


\begin{table}[t]
\caption{Performance comparison on RGB datasets  against  RGB methods for MAE metric. The lower ($\downarrow$) is the value for MAE, the better the accuracy). \textcolor{red}{Red}, \textcolor{blue}{Blue}, and \textcolor{magenta}{Magenta} colors show first, second, and third best performance.}
\captionsetup{justification=centering}
\begin{center}
\resizebox{\columnwidth}{!}{
\begin{tabular}[width=\columnwidth]{l|c|c|c|c}

\toprule

Methods  & ECSSD \cite{yan2013hierarchical} & HKU-IS\cite{li2015visual} & DUTS-TE \cite{wang2017learning} & PASCAL-S \cite{li2014secrets} \\ \hline

NLDF \cite{luo2017non}   &   .051   &  \textcolor{black}{.041}      &  \textcolor{black}{.055}     &  .083  \\

DSS \cite{hou2017deeply}   &   .051   &  \textcolor{black}{.043}      &  \textcolor{black}{.050}     &  .081  \\

BMPM \cite{zhang2018bi}   &   .044   &  \textcolor{black}{.039}      &  \textcolor{black}{.049}     &  .074  \\

Amulet \cite{zhang2017amulet}   &   .057   &  \textcolor{black}{.047}      &  \textcolor{black}{.062}     &  .095  \\

SRM \cite{wang2017stagewise}   &   .054   &  \textcolor{black}{.047}      &  \textcolor{black}{.059}     &  .085  \\

PiCANet \cite{liu2018picanet}   &   .035    &  \textcolor{magenta}{.031}    &  .040   &  \textcolor{black}{.072} \\ 

DGRL \cite{wang2018detect}   &   .043   &  \textcolor{black}{.037}      &  \textcolor{black}{.051}     &  .074  \\

CPD \cite{wu2019cascaded}   &   .037   &   .034     &  .043    &    .074  \\

EGNet \cite{zhao2019egnet}   &   .037   &  \textcolor{magenta}{.031}      &  \textcolor{magenta}{.039}     &  .080  \\

TSPOANet \cite{liu2019employing} &   .047      &  .039  &  .049      &  .082  \\

AFNet \cite{feng2019attentive}  &   .042  &  .036   &  .045   &  .076  \\

PoolNet \cite{liu2019simple}   &   .042    &  .032   &  .041   &  .076  \\

BASNet \cite{qin2019basnet}   &   .037    &  .032   &  .047   &  .083  \\

GateNet \cite{zhao2020suppress}     &   .038        &  \textcolor{magenta}{.031}        &  \textcolor{blue}{.037}        &  \textcolor{magenta}{.071}   \\

CSNet \cite{gao2020highly}  &   \textcolor{blue}{.033}   &  .      &  \textcolor{red}{.037}    &  .073  \\

LDF \cite{wei2020label}  &   .034      &  .028     &  .034   &  .067  \\

MSI-Net \cite{pang2020multi}  &   \textcolor{magenta}{.034}     &  \textcolor{red}{.029}    &  \textcolor{red}{.037}    &  \textcolor{magenta}{.071}   \\

ITSD \cite{zhou2020interactive}   &   .035     &  \textcolor{magenta}{.031}      &  .041     &  \textcolor{magenta}{.071}  \\

VST \cite{liu2021visual}   &   .037    &  .030   &  .037   &  \textcolor{red}{.067}  \\

SDC \cite{li2021looking}     &   \textcolor{magenta}{.034}         &  \textcolor{blue}{.030}    &  \textcolor{blue}{.037}  &  .111  \\

PoolNet-R+ \cite{liu2022poolnet+}     &   \textcolor{black}{.040}         &  \textcolor{black}{.034}    &  \textcolor{black}{.039}  &  \textcolor{blue}{.068}  \\

\hline
\ourmodel$_{RGB}$  & \textcolor{red}{.030}    &   \textcolor{red}{.029} & \textcolor{magenta}{.039}   &  \textcolor{blue}{.068}   \\ \bottomrule

\end{tabular}
}
\end{center}
\label{tab:RGB}
\vspace{-5mm}
\end{table}

%% file: sections/conclusion.tex
\section{Conclusion}
In this paper, we rethink RGB and RGB-D SOD from a new perspective, employing multi-level CNN and pyramidal attention features in the encoder. Our method is applicable to RGB and RGB-D SOD, where we acquire a single RGB input, estimate its depth, and fuse depth features with the RGB model to enhance saliency prediction results. Moreover, our proposed decoder architecture fuses encoder features at multi-levels and gradually upsample to predict saliency maps with refined edges and focus on the essential parts of the input image. To the best of our knowledge, we are the first to eliminate the acquisition of multiple inputs for RGB-D SOD and achieved state-of-the-art results over both RGB and RGB-D SOD datasets.

\subsubsection*{Acknowledgements}
This work was supported by Institute of Information \& communications Technology Planning \& Evaluation (IITP) grant funded by the Korea government(MSIT) $[$No. 2020-0-00062, Project Name$:$ Development of data augmentation technology by using heterogeneous data and external data integration$]$.